\documentclass[journal]{IEEEtran}

\ifCLASSINFOpdf
\else
\fi
\usepackage[table,xcdraw]{xcolor}
\usepackage[pdftex]{graphicx}

\usepackage[caption=false,font=footnotesize]{subfig}

\usepackage{cite}
\usepackage{lipsum}
\usepackage[cmex10]{amsmath}

\usepackage{tikz}
\usetikzlibrary{arrows,automata}

\begin{document}

\title{Tracking Road Users using Constraint Programming}

\author{Alexandre Pineault, Guillaume-Alexandre Bilodeau~\IEEEmembership{Member,~IEEE}, Gilles Pesant \thanks{A. Pineault, G.-A. Bilodeau and G. Pesant are with the Department
of Computer and Software Engineering, Polytechnique Montreal, Montreal,
Canada, e-mails: (alexandre.pineault@polymtl.ca, gabilodeau@polymtl.ca, gilles.pesant@polymtl.ca).}% <-this % stops a space
 %<-this % stops a space
\thanks{}}%Manuscript received ; revised }}

\markboth{}%IEEE Transactions on Intelligent Transportation Systems}%
{Pineault \MakeLowercase{\textit{et al.}}: Tracking Road Users using Constraint Programming}

\maketitle

\begin{abstract}

     In this paper, we aim at improving the tracking of road users in urban scenes. We present a constraint programming (CP) approach for the data association phase found in the tracking-by-detection paradigm of the multiple object tracking (MOT) problem. Such an approach can solve the data association problem more efficiently than graph-based methods and can handle better the combinatorial explosion occurring when multiple frames are analyzed. Because our focus is on the data association problem, our MOT method only uses simple image features, which are the center position and color of detections for each frame. Constraints are defined on these two features and on the general MOT problem. For example, we enforce color appearance preservation over trajectories and constrain the extent of motion between frames. Filtering layers are used in order to eliminate detection candidates before using CP and to remove dummy trajectories produced by the CP solver. Our proposed method was tested on a motorized vehicles tracking dataset and produces results that outperform the top methods of the UA-DETRAC benchmark.
    
\end{abstract}

\begin{IEEEkeywords}

     multiple object tracking; road users; data association; constraint programming
    
\end{IEEEkeywords}

\IEEEpeerreviewmaketitle

\section{Introduction}

\begin{figure*}
        \centering
        \includegraphics[width=0.8\paperwidth]{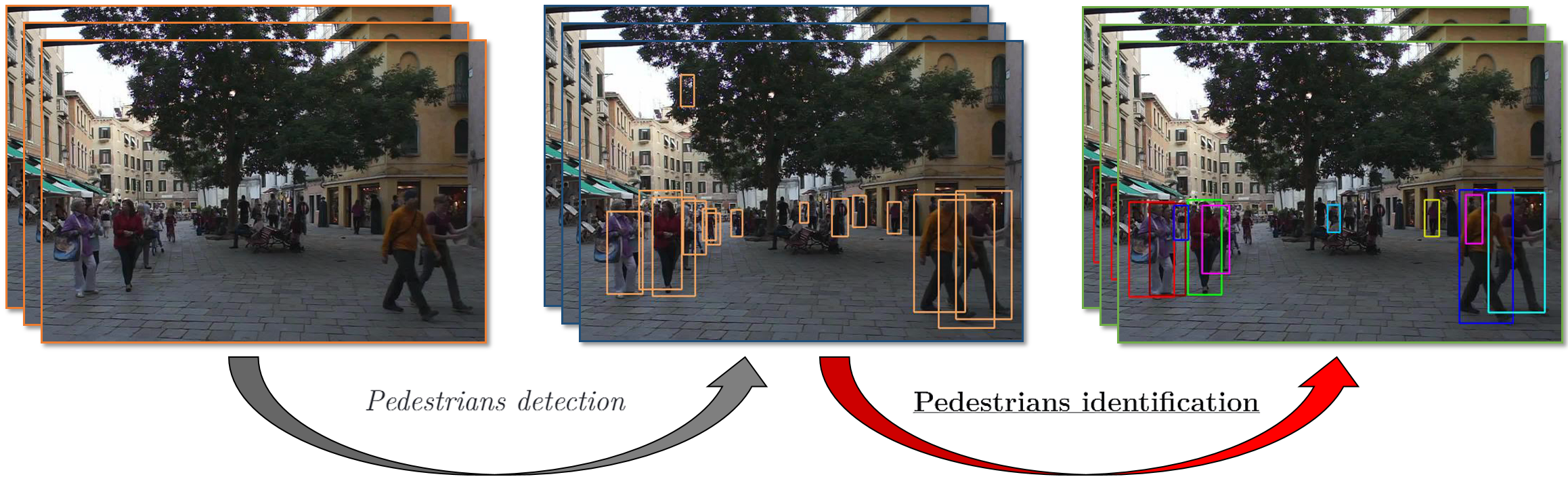}
        \caption{Solving a MOT situation involving pedestrians using the Tracking-by-detection paradigm. Pedestrians are first detected and then they are assigned a unique identifier, represented here by the various colors.}
        \label{fig:MOTdesc} 
    \end{figure*}

    \IEEEPARstart{I}{n} this paper, we address the multiple object tracking (MOT) problem in the context of traffic scenes. A very common approach to tackle the MOT  problem uses the tracking-by-detection paradigm, which divides the task into two smaller ones: the detection of objects in the video frames and the association of these detections between frames to form trajectories for the objects of interest \cite{Riahi2016,ICCV2017Sadeghian,CVPR2014Bae,CVPR2011Pirsiavash}. It is illustrated in Figure \ref{fig:MOTdesc}. The resulting trajectories can represent data about road users that is useful to solve higher-level problems such as improving security in our streets and improving traffic flows for more eco-friendly mobility. For example, tracking road users can help design more secure roads by analyzing users' behaviour and finding anomalies in their trajectories before incidents actually happen \cite{Saunier2010}.
    
\begin{figure}
        \centering
        \subfloat[]{\includegraphics[height=0.21\paperwidth]{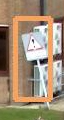}}\ \
        \subfloat[]{\includegraphics[height=0.21\paperwidth]{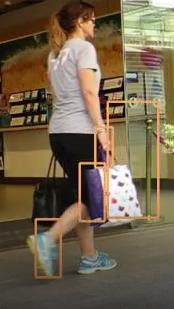}}\ \
        \subfloat[]{\includegraphics[height=0.21\paperwidth]{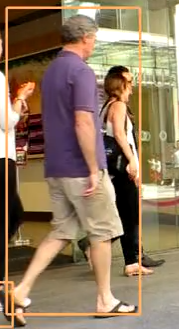}}
      \caption{Examples of detection errors. a) False detection, b) Object fragmentation, and c) Missed detection (the person behind) caused by an occlusion.}
      \label{fig:confDet} 
    \end{figure}

    This work focuses on the data association step of MOT methods to improve their performance for traffic scenes. We assume that our road user detections come from an off-the-shelf object detector. The main challenges of data association are: 1) handling occlusions which occur when a tracked entity is hidden completely or partially by other objects such as other road users or untracked scene elements (e.g. urban furniture), 2) managing incoming and outgoing objects of interest, which means that a trajectory may not last the whole video sequence, and 3) taking into account the imperfect performance of the detector, such as bounding boxes that do not bound perfectly an object, missing objects, or false detections (see Figure \ref{fig:confDet}).
    
    The data association step can be viewed as a combinatorial problem between current tracks (a track is a trajectory being built by the tracker) and new detections. For this reason, the use of constraint programming (CP) is an interesting option to explore since CP has long been used to solve various combinatorial problems \cite{1996Wallace}. As a first step, CP states a formal model of the problem to solve using finite-domain variables and constraints on them. It then explores the combinatorial space of solutions using tree search guided by variable- and value-selection heuristics and applies sophisticated inference at each node of the search tree in order to prune infeasible branches. To every variable in the model is associated a finite set called its domain: the variable can take any value in that domain. Constraints on the variables forbid certain combinations of values. Each type of constraint encapsulates its own specialized inference algorithm that filters out inconsistent values from the domains of variables. We refer the interested reader to \cite{2014Pesant} for a more detailed exposition of CP.
    
    This paper presents a MOT method that produces trajectories by matching recurring objects together using a set of constraints that use the center position and the color of detections. Our main contribution is a road user tracker that uses a CP model that is adapted to the MOT problem. Some filters are defined before and after the CP solving phase. Our method is compared with others based on data association strategies with similar simple features as the ones we are using. We show that in the case of simple features, our CP-based data association method gives better results compared with a state-of-the-art road user tracker on the UA-DETRAC benchmark. Using additional constraints, the proposed CP-based data association can be extended to include other features.

\section{Background and Related Work}
\label{review}

    \subsection{Road User MOT}

        A simple method to track road users is to use only the proximity between detections in consecutive frames to make decisions. It is possible to do so by computing the intersection over union metric that corresponds to the overlap area proportion between two bounding boxes, then choosing associations based on the highest overlap \cite{2018Bochinski}. This method has proven to be fast and was able to achieve a competitive accuracy on the UA-DETRAC benchmark. 
        
        Another approach is to use kernelized correlation filters (KCF) \cite{2015Henriques} where a filter is estimated in such a way that when applied to the region of interest to track, a strong response is produced. The goal is to find the image region that produces the most similar response in the subsequent frames. This technique was combined with a background subtraction process to track road users \cite{CRV2016Yang}. Although successful, this method lacks a proper data association process to better handle occlusions. A Kalman filter is also often used to help association tof road users by predicting missing detections \cite{Ooi2018}.  Another work\cite{2016Jodoin} uses background subtraction, and manages occlusion situations by considering vehicles travelling together in only one track. A state machine manages the combination of tracklets when road users merge together or choose different paths. It uses keypoints as an appearance model and does data association only on two consecutive frames. Finally, another approach is to track keypoints using optical flow, and grouping these using a common motion criteria \cite{CRV2006Saunier}. The shortcoming of this method is that objects are not localised very well because it is based on sparse keypoints. Furthermore, keypoint detection and tracking rely on motion. Stopped road users are not tracked. 
        
    \subsection{General MOT}

        State-of-the-art performance in multiple object tracking is currently obtained with methods using deep neural networks to extract many automatically-learned features \cite{ICIP2017Chen,ICCV2017Sadeghian}. One of these methods \cite{ICIP2017Chen} is using a convolutional neural network (CNN) to build two classifiers: one for separating object categories and another for instance classification which will differentiate objects of a same category. In this approach, the previous classifiers are connected to a particle filter which, combined with the appearance model, makes a prediction of where the object will appear next. The idea of using object categories in tracking was also exploited in the work of Ooi et al. \cite{Ooi2018} in road user tracking.  More recently, re-identification features are gaining popularity to describe an object of interest in multiple object tracking situations \cite{ICCV2019Bergmann, CVPR2019Feng}.  
        
        Another paper using recurrent neural networks (RNNs) has focused more on interactions between objects in order to disambiguate matches \cite{ICCV2017Sadeghian}. A spatio-temporal appearance model, where a CNN takes bounding boxes as input and outputs features to a LSTM capable of memorizing long-term feature dependencies, is added to a motion feature extractor and an interaction feature extractor. The last part is made of a grid representing where nearby detected objects are located and a second LSTM module.
        
        There are more and more methods that start using visual object tracking (VOT) techniques in a MOT context. In the VOT problem, there is only one given detection in the first frame, that states which object is to be tracked. Thus, VOT method uses strategies to differentiate the foreground from the background. This approach is problematic in MOT situations since there are many similar foreground objects and it becomes difficult to ensure that all objects are tracked since two trackers initialized on two different objects may end up following the same after a few frames. A possible way to mitigate the risk of these situations is to use a VOT for each object to make predictions about the position of the object in the next frame. Then these predictions can be incorporated in a data association method, where they are considered in a more global context \cite{CVPR2019Chu, CVPR2019Feng}.
        
        These state-of-the-art methods focus mainly in correcting missing and spurious detections using a prediction method and defining robust features for data association. They do not study how to make the data association itself. They show that a strong appearance model and predicting where an object should appear next to filter object detections are two important components of a MOT system. However they use a limited data association strategy, as data association decisions are not taken over several frames by combining several observations of the objects in a video. Occlusions are better resolved over several frames. Therefore, a more robust data association method is always desirable.
        
        Typical data association methods used in tracking are the Hungarian algorithm, the joint probabilistic data-association filter (JPDAF), and the minimum-cost flow algorithm. The Hungarian algorithm finds a maximum cost matching in a bipartite graph where the costs are on edges \cite{Kuhn1955}. As the assignment problem can be formulated as a bipartite graph, this algorithm is one way to get the solution. It was used in several works. For example, the Hungarian algorithm can be combined with tracklets (incomplete track parts) to make the associations \cite{CVPR2014Bae}. In this work, the tracklets are ranked according to a confidence metric considering occlusions, number of frames covered and how well the detections fit. Online detections are matched to existing tracklets with high confidence using the Hungarian algorithm maximizing the affinity level. The next step is to globally match these new associations with low level detections using once again the Hungarian algorithm. 
        
        JPDAF is a statistical method that can track objects based on what is the most likely outcome for each trajectory. It considers any detection available, but also the possibility of a missing object or a false detection. It was used several times for the MOT problem, for example, in the work of \cite{Rezatofighi2015} and \cite{CSM2009Shalom}.
        
        The min-cost flow algorithm combines a cost function with a greedy algorithm in order to obtain the required tracking associations \cite{CVPR2011Pirsiavash}. The goal here is to formulate the data association as a minimum-cost flow problem, then to compute shortest paths on the flow network with detections at each frame as nodes, from the first appearance of an object to its last appearance in the scene. This process helps ensure that the resulting trajectories are as smooth as possible.

\section{CP Background}

    In this section we present some useful background about constraint programming. There are three high level constraints used by our approach:
    \begin{enumerate}
    
        \item \texttt{AllDifferent}: Applied to a set of CP variables $\{ x_1, x_2, \ldots, x_n \}$, this constraint states that it is not permitted that two variables $x_i$ and $x_j$ share the same value:
        
        \begin{equation}
            x_i \neq x_j  \qquad \forall \;
            1 \leq i < j \leq n
        \end{equation}
        
        \item \texttt{Inverse}: Given two arrays of CP variables $\langle x_1, x_2, \ldots, x_m \rangle$ and $\langle y_1, y_2, \ldots, y_n \rangle$, the following equations must hold:
        % GCC: https://sofdem.github.io/gccat/gccat/Cinverse.html
        \begin{equation}
            x_i \in [1, n] \Rightarrow y_{x_i} = i \qquad \forall\; 1 \leq i \leq m
        \end{equation}
        \begin{equation}
            y_j \in [1, m] \Rightarrow x_{y_j} = j \qquad \forall\; 1 \leq j \leq n
        \end{equation}
        This constraint is useful to maintain consistency between dual representations.
           
        \item \texttt{Regular} and \texttt{CostRegular}: These express a constraint as membership to a regular language. Given a sequence of variables $\langle x_1, x_2, \ldots, x_n \rangle$ and an automaton $\cal A$, each solution to the constraint corresponds to an assignment of values to these variables that spell out a word recognized by $\cal A$~\cite{CP2004Pesant}. \texttt{CostRegular} is the optimization version of the latter constraint: it takes as additional parameters the costs associated with each combination of variable, value, and automaton state, as well as a cost variable equal to the sum of individual costs in a solution~\cite{Demassey2006}.
        
    \end{enumerate}
    
    The inference algorithm associated with each of these constraints removes every inconsistent value in the domain of each variable.

\section{Methodology}
\label{metho}

\begin{figure*}
        \centering
        \includegraphics[width=\linewidth]{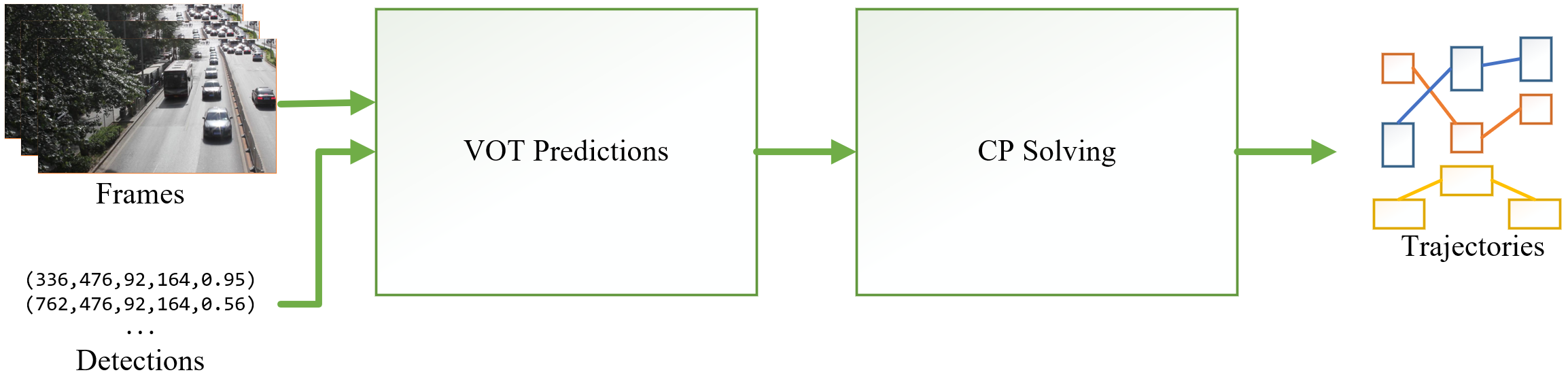}
        \caption{Diagram showing the  interactions of the high-level modules of our method.}
        \label{fig:meth-diagram}
    \end{figure*}

    Our road users tracking approach can be summarized as follows. First, detections are obtained for every video frame using a road user detector. A VOT is also applied to predict the position of the objects detected in the previous frame, in the new one. These detections are then filtered based on their confidence and redundancy. Detections can be redundant if a predicted box matches a detection in the bouding boxes initial set. The predicted bounding boxes should be ignored in this case. Then, the detections are used to instantiate the CP model and the model is solved by forming tracks with the detections. Finally, unused tracks are removed before outputting the final result (a set of trajectories). Our method is detailed in the following and a high-level view is shown on Figure \ref{fig:meth-diagram}.
    
    \subsection{Constraint Programming Model}
    
        \subsubsection{Main Variables}
            
            In order to associate tracks to detections, variables
            \begin{alignat}{2}
                t_{ij} \in \{1,2,\ldots,\tau\} & \qquad & \forall \; 1 \leq i \leq m,\; 1 \leq j \leq n_i
                \label{eq:trajDef}
            \end{alignat}
            identify to which track (out of $\tau$ available ones) detection $j$ at frame $i$ belongs, where $m$ is the number of frames and $n_i$ the number of detections at frame $i$.

            Such a representation is convenient to avoid unnecessary symmetries since it uses the smallest number of possible variables: one per detection. However, referring to consecutive detections of a given track is not possible before we obtain the solution; there is no way to know the indices of the variables that will be part of a same track before the solving step. 
            Since we need this to express some of our constraints, a second array of variables is defined,
            \begin{alignat}{2}
                d_{ik} \in \{1,2,\ldots,\tau\} & \qquad & \forall \; 1 \leq i \leq m,\; 1 \leq k \leq \tau
                \label{eq:detDef}
            \end{alignat}
            identifying the detection assigned to track $k$ at frame $i$.
            Whenever $\tau > n_i$ for some frame $i$, values greater than $n_i$ do not represent actual detections --- they are however necessary because we will require that detections be uniquely associated to tracks (see Equations~\ref{eq:alldiff_t} and \ref{eq:alldiff_d}).
            
            To ensure that these two dual representations remain consistent with each other, we link them using an \texttt{inverse} constraint
            \begin{alignat}{2}
                d_{i\star} = \texttt{inverse}(t_{i\star}) & \qquad & \forall\; 1 \leq i \leq m 
                \label{eq:cstrInv}
            \end{alignat}
            relating each variable to its counterpart in the other array. the star symbol ($\star$) is used to indicate the selection of a subarray.
        
        \subsubsection{Frame Consistency}
        
             The next step is to ensure that it is impossible to find a specific track or detection more than once at a given frame. This is a common requirement for which CP provides an \texttt{AllDifferent} constraint. Such constraints have been specified for this purpose using the same variables as the previous equations.
            
            \begin{alignat}{2}
                \texttt{AllDifferent}(t_{i\star}) & \qquad & \forall\; 1 \leq i \leq m
                \label{eq:alldiff_t}\\
                \texttt{AllDifferent}(d_{i\star}) && \forall\; 1 \leq i \leq m
                \label{eq:alldiff_d}
            \end{alignat}
        
        \subsubsection{Position Constraints}
            
            After the model structure is set, constraints on features such as position can be considered. First, an array of integers is created for each dimension in a frame ($x$ and $y$ coordinates). These arrays, indexed by frame and then by detection ID, contain the center of each detection bounding box. It is then possible to formulate a constraint restricting the distance between two consecutive detections in a track by stating that the one-dimensional distance along the $X$ and the $Y$ axes separating them may not be greater than thresholds $\lambda_x$ and $\lambda_y$ :
            
            \begin{alignat}{2}
                |x_{i,d_{ik}} - x_{{i+1,d_{i+1,k}}}| \leq \lambda_x & \qquad & \mkern-18mu & \forall\; 1 \leq i < m, 1 \leq k \leq \tau\\
                |y_{i,d_{ik}} - y_{{i+1,d_{i+1,k}}}| \leq \lambda_y && \mkern-18mu & \forall\; 1 \leq i < m, 1 \leq k \leq \tau 
            \end{alignat}
            
            The same process has been tested using object scale indicators (width and height) but this feature did not help improve the solution.
         
        \subsubsection{Appearance Model}
        
            An appearance model is useful for minimizing the risk of mismatch between two nearby objects. Our proposed appearance model assigns a color class label to each detection available. The color classes are obtained by clustering observed object color histograms in several videos. Clustering is performed with K-Means using the Bhattacharya distance between the two color histograms. The learned classes are then used during tracking. A detection in a frame is assigned the color class label of the nearest cluster.  

            A \texttt{CostRegular} constraint is applied to each track preventing the association of two objects with contrasting colors. Possible transitions between color classes for a given object at each frame is governed by a state machine. Transitions between similar color classes are allowed but a cost is applied based on the Bhattacharya distance between the class centers. The solver will use the sum of all these costs as a minimization objective.
            
            \begin{gather}
                c_{ik} = \texttt{colour}(d_{ik})\\
                \texttt{CostRegular}(c_{\star k}, {\cal A}, C, a_{k}) \qquad \forall\; 1 \leq k \leq \tau\\
                \texttt{min}\left(\sum_{k=1}^{\tau} a_{k}\right)
            \end{gather}
            
            In these equations, $c_{ik}$ represents the list of color class labels for the track detections acting as transition variables between the automaton's  states. $\cal A$ and $C$ correspond respectively to the automaton and the cost matrix for every possible transition. Variable  $a_{k}$ represents Track $k$'s total cost computed by the regular constraint.
            
\begin{figure}
                \centering
                \begin{tikzpicture}[->,>=stealth', shorten >=1pt, auto, node distance=2.5cm, semithick, every node/.style={scale=1.1}]
                    \tikzstyle{every state}=[draw=none,text=white]
            
                    \node[initial,state, fill=black]                (A)              {\large $i$};
                    \node[state, fill=orange]                       (B) [below of=A] {\large $o_v$};
                    \node[state, fill=orange]                       (C) [below of=B] {\large $o_c$};
                    \node[state, fill=black!15!yellow, text=black]  (D) [right of=B] {\large $y_v$};
                    \node[state, fill=black!15!yellow, text=black]  (E) [right of=C] {\large $y_c$};
                    \node[state, fill=black!12!red]                 (F) [left of=B]  {\large $r_v$};
                    \node[state, fill=black!12!red]                 (R) [left of=C]  {\large $r_c$};
                    
                    \path   (A) edge [loop right]                   node                        {E} (A) 
                                edge                                node [pos=0.4]              {O} (B)
                                edge                                node [pos=0.56, above=3pt]  {Y} (D)
                                edge                                node [pos=0.56, above=3pt]  {R} (F)
                            (B) edge [in=65, out=35, looseness=8]   node [pos=0.5, above]       {O} (B)
                                edge [bend left=15]                 node                        {E} (C)
                                edge [bend left=15]                 node                        {Y} (D)
                                edge [bend left=15]                 node                        {R} (F)
                            (C) edge [bend left=15]                 node                        {O} (B)
                                edge [loop right]                   node                        {E} (C)
                                edge                                node [pos=0.3, below=3pt]   {Y} (D)
                                edge                                node [pos=0.3, below=3pt]   {R} (F)
                            (D) edge [bend left=15]                 node                        {O} (B)
                                edge [loop right]                   node                        {Y} (D)
                                edge [bend left=15]                 node                        {E} (E)
                            (E) edge                                node [pos=0.3, below=3pt]   {O} (B)
                                edge [bend left=15]                 node                        {Y} (D)
                                edge [loop right]                   node                        {E} (E)
                            (F) edge [bend left=15]                 node                        {O} (B)
                                edge [loop left]                    node                        {R} (F)
                                edge [bend left=15]                 node                        {E} (R)
                            (R) edge                                node [pos=0.3, below=3pt]   {O} (B)
                                edge [bend left=15]                 node                        {R} (F)
                                edge [loop left]                    node                        {E} (R);
                \end{tikzpicture}
                \caption{Small state machine example illustrating the doubled states: visible (v) and occlusion (c) for three color classes: red (d), orange (o) and yellow (y). Transition values correspond to the detection color (R, O, Y) or empty (E) when there is no detection.}
                \label{fig:states}
            \end{figure}
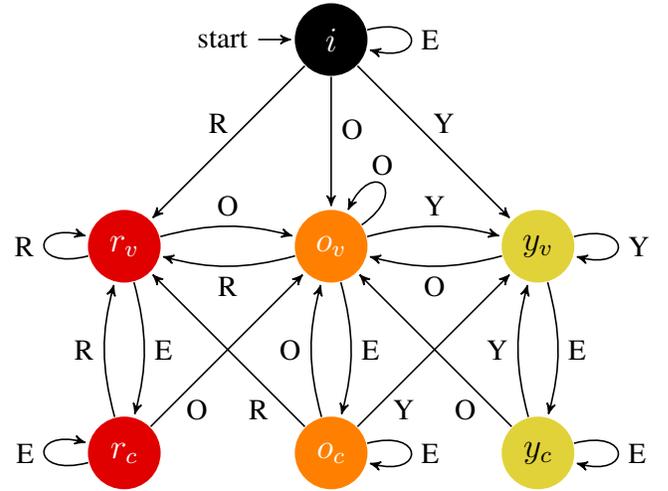

            Each color class is represented by two states, depending on whether the object is currently being occluded or not (e.g. red object and red occluded object). Using different states for occlusions allows to set a cost for transitions from a valid detection to an absence of detection while preserving the object appearance even when it is occluded for an extended duration (see Figure \ref{fig:states}). The cost are based on the Bhattacharya distance between each color class histograms.
            
    \subsection{Solving Strategy}

        \subsubsection{Variable Selection}

            As stated in the introduction, CP explores the combinatorial space of solutions by branching in a search tree that enumerates all possible solutions. To solve our model, we first branch on the $t_{ij}$ variables since it is the main array containing the smallest number of variables necessary to specify a solution whereas in the case of the $d_{ik}$ variables, all possible tracks are instantiated in case they are needed. To allow the state machine to handle occlusions or the absence of a detection for a given track, it is necessary to branch on the $c_{ij}$ variables. The lexicographic order specified will make sure that branching will always at first be made on $t_{ij}$ variables, then on $c_{ij}$ variables. Without including $c_{ij}$ variables in the branching order, some subconstraints of the \emph{CostRegular} might be ignored in the event of occlusions (no $t_{ij}$ linked to the $c_{ij}$). In both cases, variables are considered frame by frame, then detection identifier by detection identifier in ascending order.
            
        \subsubsection{Value Selection}
        
            Since the $t_{ij}$ variables are fixed by the solver from the first to the last frame, the use of prediction becomes an interesting option to guide the search. By computing every distance between possible pairs of consecutive detections, it is possible to obtain a ranked list that the solver can use to branch on closest detections first for each $t_{ij}$ except for the last frame. This way, we minimize the distance between each consecutive pair of objects without altering the model objective oriented towards the appearance preservation.
            
            The inherent symmetries are addressed in the branching to control the number of opened tracks available as candidate for $t_{ij}$ variables. The domain of each of these variables is reduced to all previously used tracked index plus one other that may be opened if necessary. This strategy was applied before to the Steel Mill Slab Design Problem \cite{CPAIOR2008VH}.
    
    \subsection{Pre Solving Computations}
    \label{ssec:Presolve}
    
        As mentioned in the introduction, detectors are prone to miss objects of interest. To compensate this weakness, we developed a pretracking method using an VOT to add new detections, using this process:
        \begin{enumerate}
            \item  We initialize a KCF tracker \cite{2015Henriques} for each given detection of the first frame. With KCF, the image is correlated with a filter that is learned for each road user. The object is localized based on the strength of the filter response. It is a very fast tracker. 
            \item  For the next frames, one by one, all KCF trackers make a prediction about the next bounding box of the object. For each prediction that is not redundant with a detection of the current frame, the bounding box is kept. For all detections in the current frame that do not match a prediction, a new tracker is initialized.
            \item Each created tracker has a lifespan that allows it to make predictions for a fixed small number of frames. It is assumed that the tracker is reliable only on a small time window. To avoid adding detections based on a false detection, predicted boxes are only added to the existing detections if there is a match between an existing detection and the tracker prediction in a subsequent frame.
            
        \end{enumerate}
        
        \begin{figure}
            \centering
            \includegraphics[width=0.4\paperwidth]{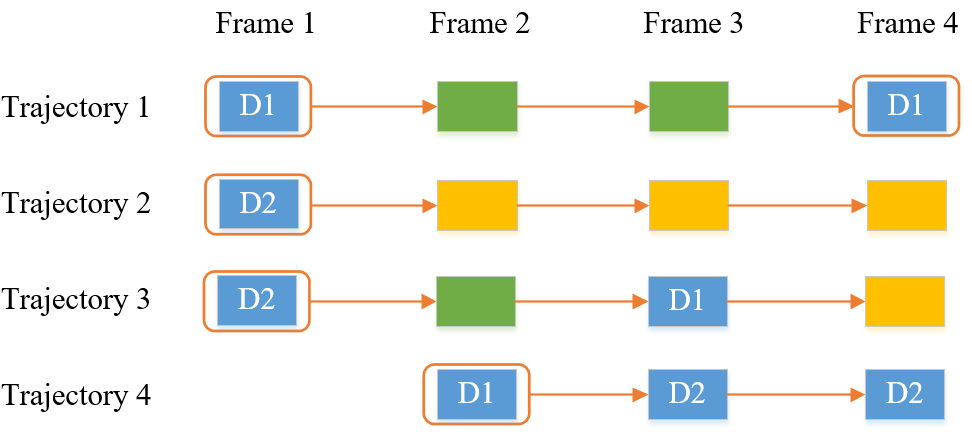}
            \caption{Illustration of the detection prediction process where blue boxes are detections from an object detector, yellow and green boxes are respectively discarded and added detections using KCF. The orange frames indicate the initialization of a tracker.}
            \label{fig:DetPred}
        \end{figure}
        
        Figure \ref{fig:DetPred} illustrate many possible situations. For trajectory 1, a tracker initialized in the the first frame add two detections (frames 2 and 3). This is the ideal way to manage a two frames occlusion; two boxes are added when the object is hidden, then the tracker response match again the object and reinitialized itself since is prediction life is over. Trajectories 2 and 3 illustrate cases where we assume that the tracker is mistaken. If a VOT tracker is initialized on a false detection, no predicted boxes should be kept (trajectory 2). For the third trajectory, the tracker is able to match a real detection, but produces afterward another detection that remains unmatched. To avoid the addition of too many detections that would fall into the false positives category, the last detection is simply discarded.
    
    \subsection{Post Solving Computations}
    \label{ssec:Postsolve}
    
        Once the solver has completed the associations, another type of filtering is required due to how the model is built. The model is unable to eliminate detections entirely --- the only choice that it has is to put each of them in dummy tracks, which will have to be deleted afterward. To delete them, a filter removes every track that contains fewer than $\beta_D$ detections from the object detector, a parameter set to a fixed value for all sequences.

        After the suppression of the unwanted tracks, it is possible to fill gaps in the remaining tracks. A small state machine searches for places where there are $\gamma_D$ or less consecutive missing detections and fills these empty spots by adding new detections using linear interpolation to find the correct center position and size of the bounding box for each one. This has to be done carefully because a too large gap probably means that an identity change has happened. Therefore, filling missing detections in these situations would probably mean that false detections are added.
        
    \subsection{Batch Resolution Process}
    \label{ssec:A-Batch}
    
        Since this model uses optimization to find the best solution to the MOT problem applied to large instances, the time required to find a solution is much longer than the one obtained by simply solving the satisfaction problem. Moreover, the number of computations required to find the optimal solution  grows exponentially in the size of the problem instance which makes the minimization process intractable to complete in practice.
        
        This situation was addressed by dividing a video sequence into multiple batches of $\kappa$ frames. Except for the first batch, to insure that the transition between batches stay as smooth as possible, the first $\beta$ frames of the trajectories were taken from the end of the previous batch. Frames containing no detection were ignored by this batch system, which means that a batch is not always constituted of consecutive frames. This is also valid for the overlapping section of the batch. If a sequence would contain too many empty consecutive frames, the next batch will start at the first frame containing at least a detection and will be be completely independent of the previous one. The values that gave the best trade-off between accuracy of the global solution and computation time was 30 for $\kappa$ and 5 for $\beta$.

\section{Experimental Setup}

    \subsection{Evaluation Metrics}
    
        To evaluate our method, we use the standard MOT metrics that account for three types of errors at each frame $i$: 1) Identity switches ($IDS_i$) or the number of mismatches, 2) the missing detections or false negatives ($FN_i$) which are the number of objects that are not tracked, and 3) the false positives ($FP_i$) which stands for the number of detections representing no object of interest. With these values, it is possible to compute the \textit{MOTA} score \cite{EURASIP2008Bernardin}

        \begin{equation}
            \textit{MOTA} = 1 - \dfrac{\sum_{i} ( FP_i + FN_i + IDS_i)}{\sum_{i} N_i}
            \label{eq:MOTA}
        \end{equation}

        where $N_i$ is the total number of ground-truth objects per frame and it is summed throughout the entire video. The matching between tracking results and ground truth is done by measuring the intersection over union (IoU) of the two boudning boxes. If the overlapped area is higher than a certain percentage, the match is considered valid.

        A second metric known as \textit{IDF1} balances the recall and precision for each trajectory into a single value. This is done by dividing the total number of accurately matched detections to the average number of ground truth and given detections \cite{ECCVW2016Ristani}. Finally, mostly tracked trajectories ($MT$) is the ratio of ground truth that possess a matching hypothesis for at least 80\% of their life span and mostly lost trajectories ($ML$) is the number of times this ratio is under 20\%. Fragmentation ($Frag$) indicates how many times any trajectory got interrupted over its length.
        
        %Finally, some benchmark more specialized in object detections have combined the MOTA to the PR-curve evaluating the performance of detectors to define the PR-MOTA. To compute this metric, many confidence thresholds are chosen. For each threshold value, the tracking method is executed using only detections with a higher confidence percentage. The result of this process is a 3D curve where each MOTA value is represented alongside the PR-curve \cite{2015Wen}.

    \subsection{Dataset}
    
        We tested our method on the UA-DETRAC tracking challenge \cite{2015Wen}. This dataset contains 60 training sequences and 40 for the testing set. Only motorized vehicles such as cars, buses and trucks are tracked. We did not submit our results on the benchmark due to technical reasons (challenge website down, evaluation toolkit not cross-platform) to obtain the PR metrics. Instead, the code of the second best methods (IOU tracker) \cite{2018Bochinski} was downloaded, executed and tested using the usual MOTA metric (the code of the best method was not functionnal. Besides, it was only marginally better than the second one). Detections were filtered with the same parameters for both our method and the competing one. We used the same source detection from R-CNN.
    
        We used a second dataset: the 2D MOT 2015 challenge \cite{MOTChallenge2015} to discover if our method could be extended from vehicules to pedestrian tracking. This dataset provides twenty-two video sequences divided equally in a training set and a testing set. There are many additional difficulties compared to UA-DETRAC such as moving cameras, variable frame rates (sometimes 7-10 FPS) and less elevated camera angles that are more prone to occlusions.
    
    \subsection{Constraint Programming Solver}
    
        The constraint programming solver used in this paper is IBM ILOG CP (version 1.6) augmented with our own implementation of the \texttt{CostRegular} constraint. %All developed code is available on GitHub\footnote{Link will be added upon acceptance}. % TODO: link

\section{Results}
    
     Three series of experiments were conducted to evaluate how well our proposed method performs compared to others. First, the quality of our tracking approach is compared to a state-of-the-art method on the UA-DETRAC dataset. Second, we extend our method to the 2D MOT 2015 pedestrian dataset. Third, an ablation study evaluates the isolated performance of the components of our method.
    
    For all experiments, the minimal overlap value between the detection and the object areas for true positive was fixed to 50\% in the MOTA computation. No parameter optimization was done using sequences from any testing sets.
    
    \subsection{UA-DETRAC}
    
         To make sure that the way to compare the methods is fair, the same detection confidence parameter was used every time. Since the post-solving computations were not improving results for UA-DETRAC sequences, it was turned off for the results of this section.

        \begin{table*}[htb]
            \centering
            \caption{Results on the UA-DETRAC test dataset.  Bold indicates best MOTA result. The first category group (Easy, Medium and Hard) differentiates sequences according to the difficulty level while the second category group (Sunny, Cloudy, Rainy and Night) is about illumination and weather conditions. All groups of sequences are mutually exclusive in their category, but a single sequence can be found in multiple categories. Higher MOTA is better as are lower IDS, FN and FP}
            \begin{tabular}{r|rrrr|rrrr}
            \hline
            Sequences & \multicolumn{4}{c|}{CP + R-CNN}                           & \multicolumn{4}{c}{IOU + R-CNN}                          \\
            (per category)    & \textit{MOTA}  & \textit{IDS} & \textit{FN} & \textit{FP} & \textit{MOTA} & \textit{IDS} & \textit{FN} & \textit{FP} \\ \hline
            Easy      & \textbf{64.39} & 3840         & 31675       & 6833        & 60.66         & 103          & 40824       & 5861        \\
            Medium    & \textbf{63.70} & 11533        & 74538       & 15159       & 58.36         & 310          & 98117       & 17683       \\
            Hard      & \textbf{48.38} & 10316        & 126451      & 6704        & 45.58         & 215          & 142372      & 8686        \\ \hline
            Sunny     & \textbf{90.65} & 1748         & 21118       & 3131        & 64.09         & 35           & 24348       & 2980        \\
            Cloudy    & \textbf{76.42} & 6392         & 51383       & 7766        & 69.04         & 221          & 61163       & 7784        \\
            Rainy     & \textbf{58.42} & 9961         & 94828       & 10793       & 35.63         & 126          & 112843      & 13254       \\
            Night     & \textbf{71.24} & 7588         & 65335       & 7006        & 41.98         & 246          & 82959       & 8212        \\ \hline
            Global    & \textbf{57.52} & 25689        & 232664      & 28696       & 53.51         & 628          & 281313      & 32230       \\ \hline
            \end{tabular}
            \label{tab:Res-1-DETRAC}
        \end{table*}
        
        Table \ref{tab:Res-1-DETRAC} present the tracking results obtained by our method (CP) for all sequences of the test set. The MOTA was computed globally by considering the total number of errors and objects. The R-CNN public detections were used for both the IOU tracking method and ours (CP). These results show that our method is able to outperform IOU in every situations and globally. The MOTA is affected by the difficulty level, especially by the hard sequences where the false negative errors are higher than in any other sequence groups. Weather conditions and exterior luminosity also impact tracking performance. While tracking tends to be easier in sunny weather (highest results), rain looks like the most difficult environmental factor to deal with.
        
        Our strategy that helps to improve the quality of detections seem to have a high impact on the global results since the number of false positive and false negative errors are lower than those obtained by the IOU method for the same detections. However the number of ID switches is significantly higher. The main reason explaining this situation comes from the CP model itself. While the objective of the \texttt{CostRegular} constraint includes the minimization of the number of occlusions found in open trajectories, the branching strategy always tries to use already open trajectories first. Thus, the solver will implicitly prioritize compact solutions. Therefore, while our solution reduces significantly false positives and false negatives by handling better occlusion situations, it may result in a higher number of ID switches. Note that even if the number of available trajectories is determined before the resolution process, we made sure that there was always at least one unused trajectory.
        
        In order to give more details about the performance of our method, the results obtained on a subset of the testing set are presented in Table \ref{tab:Res-2-DETRAC}. The five best and worse sequences in term of MOTA are presented. The distribution of errors made by the two compared methods is still similar to what was observed in the global results, especially regarding the number of IDS and FN mistakes.
    
        \begin{table*}[htb]
            \centering
            \caption{Detailed results on the sequences in which we obtained the five highest and lowest MOTA values (UA-DETRAC, testing set). Bolface means best MOTA result. Higher MOTA is better as are lower IDS, FN and FP}
            \begin{tabular}{l|rrrr|rrrr}
            \hline
            \multicolumn{1}{r|}{Sequences} & \multicolumn{4}{c|}{CP + R-CNN}                           & \multicolumn{4}{c}{IOU + R-CNN}                           \\
            \multicolumn{1}{r|}{(Name)}    & \textit{MOTA}  & \textit{IDS} & \textit{FN} & \textit{FP} & \textit{MOTA}  & \textit{IDS} & \textit{FN} & \textit{FP} \\ \hline
            MVI\_40712                     & \textbf{83.97} & 449          & 2726        & 613         & 82.49          & 30           & 3763        & 336         \\
            MVI\_39211                     & \textbf{78.96} & 28           & 846         & 37          & 78.00          & 0            & 866         & 51          \\
            MVI\_39051                     & \textbf{77.98} & 71           & 306         & 161         & 75.62          & 2            & 471         & 119         \\
            MVI\_40854                     & 77.24          & 485          & 3222        & 867         & \textbf{78.57} & 10           & 3835        & 460         \\
            MVI\_39271                     & \textbf{76.50} & 189          & 1350        & 459         & 74.17          & 1            & 1710        & 477         \\ \hline
            MVI\_40763                     & \textbf{36.50} & 309          & 5971        & 33          & 30.62          & 3            & 6741        & 115         \\
            MVI\_40792                     & \textbf{33.40} & 537          & 7044        & 470         & 29.60          & 3            & 8161        & 332         \\
            MVI\_40863                     & \textbf{31.83} & 937          & 20966       & 392         & 29.43          & 6            & 21370       & 525         \\
            MVI\_39501                     & 30.51          & 140          & 2973        & 902         & \textbf{33.62} & 4            & 3118        & 687         \\
            MVI\_40761                     & \textbf{22.22} & 629          & 15655       & 36          & 18.93          & 4            & 16839       & 117         \\ \hline
            \end{tabular}
            \label{tab:Res-2-DETRAC}
        \end{table*}
        
\begin{figure}
            \centering
            \subfloat[MVI\_40712]{\includegraphics[width=0.18\paperwidth]{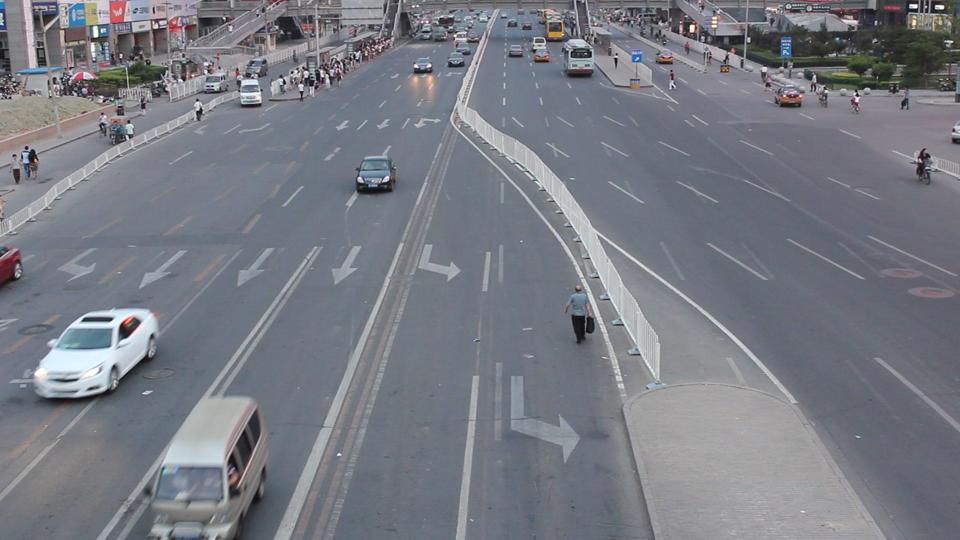}}\ \
            \subfloat[MVI\_39211]{\includegraphics[width=0.18\paperwidth]{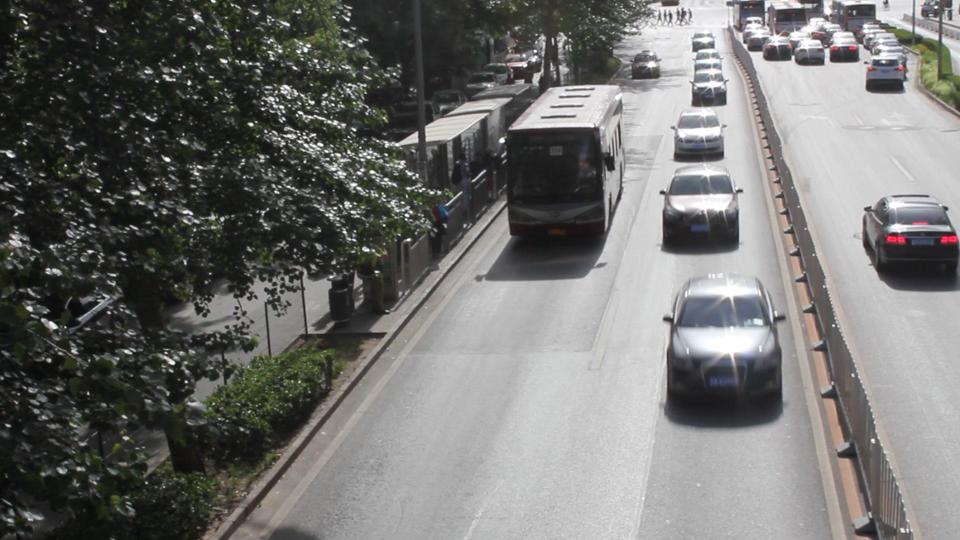}}\ \
            \subfloat[MVI\_39051]{\includegraphics[width=0.18\paperwidth]{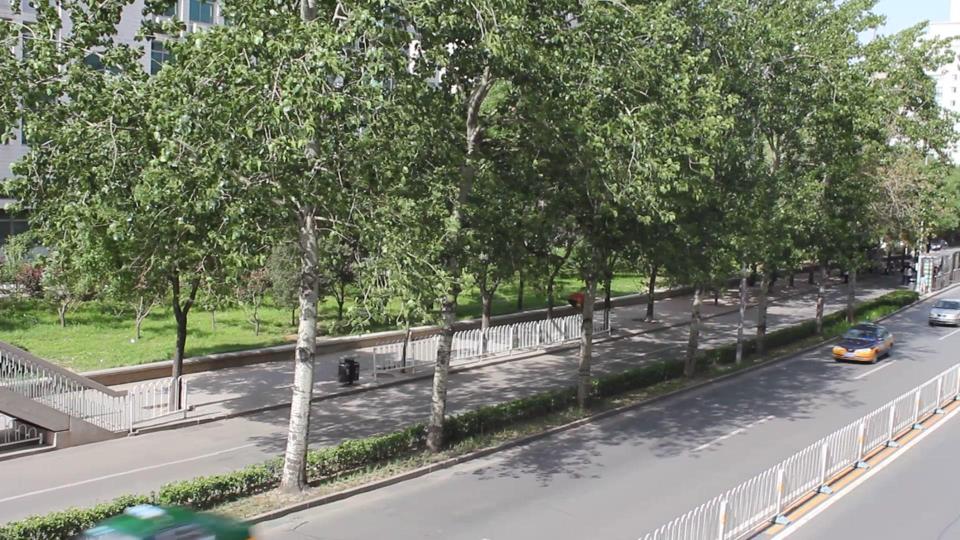}}\ \
            \subfloat[MVI\_40854]{\includegraphics[width=0.18\paperwidth]{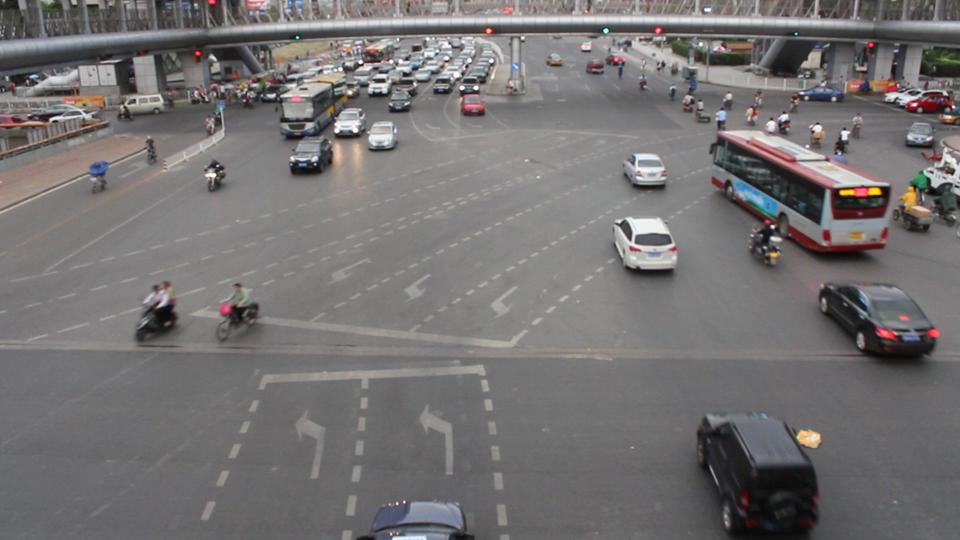}}\
            
            \subfloat[MVI\_40761]{\includegraphics[width=0.18\paperwidth]{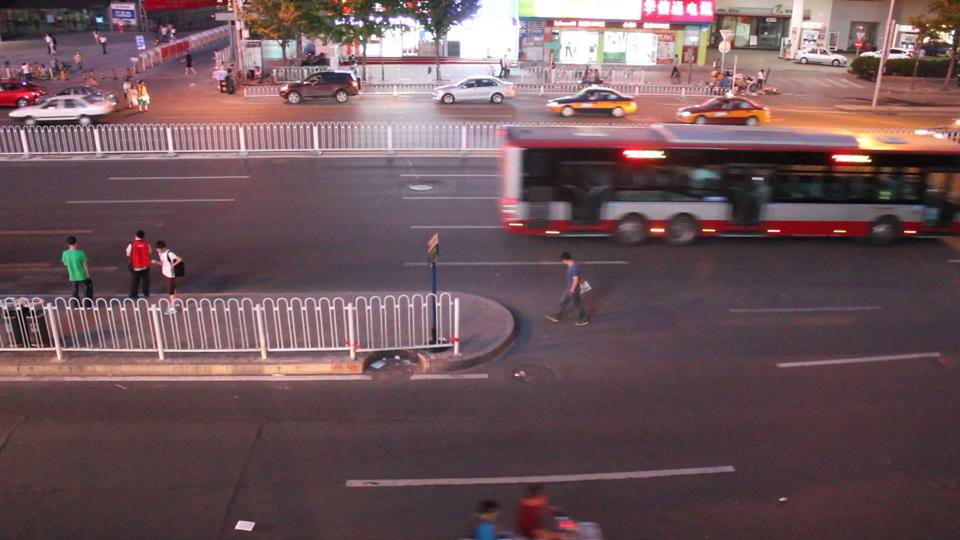}}\ \
            \subfloat[MVI\_39501]{\includegraphics[width=0.18\paperwidth]{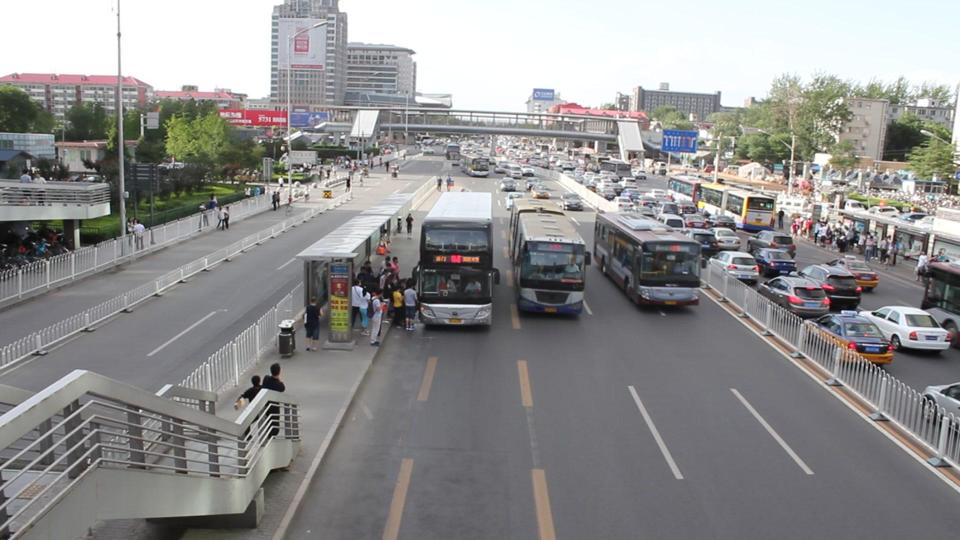}}\ \
            \subfloat[MVI\_40863]{\includegraphics[width=0.18\paperwidth]{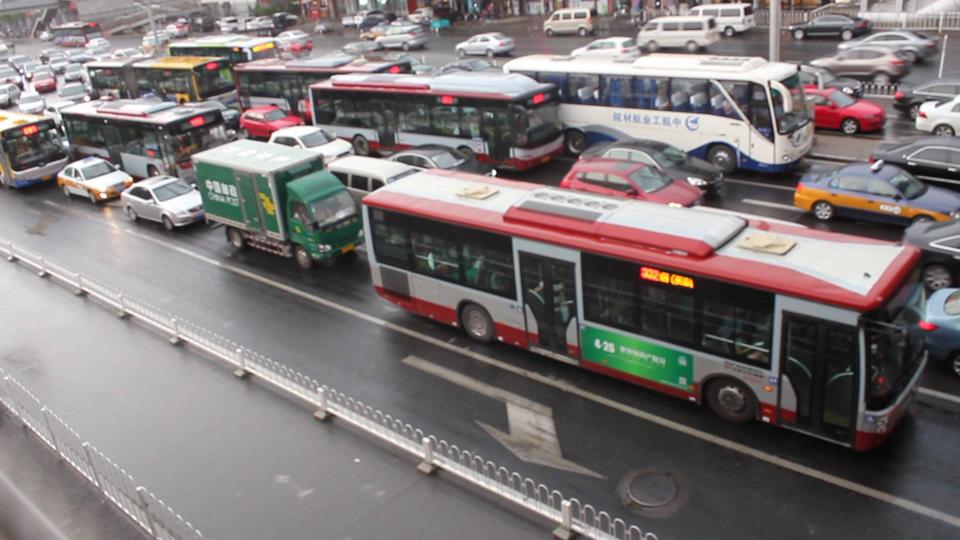}}\ \
            \subfloat[MVI\_40792]{\includegraphics[width=0.18\paperwidth]{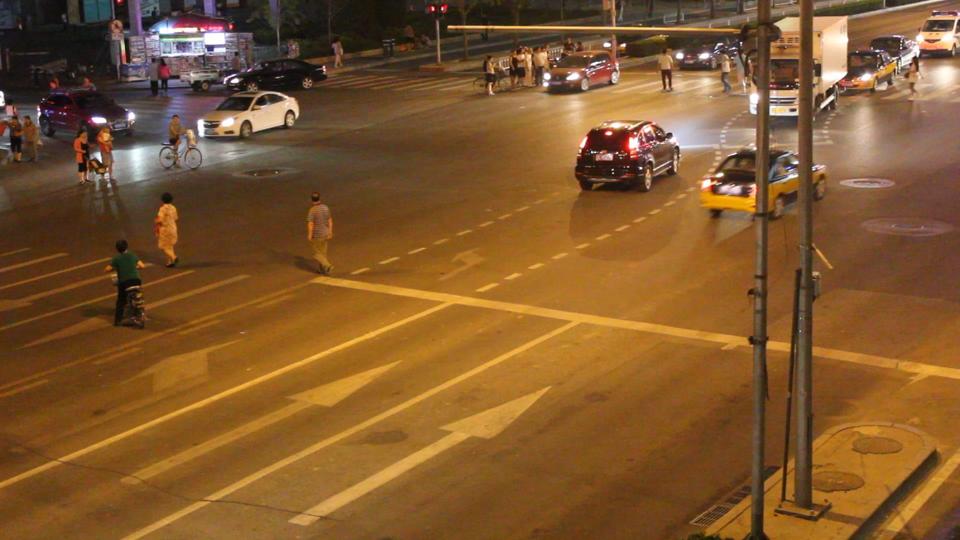}}\
            \caption{Filming perspective for different video sequences on UA-DETRAC sequences. The first row contains sequences in which our method performed the best while the second row contains the worst ones.}
            \label{fig:DETRAC-img}
        \end{figure}

        To understand why there is an important variability between sequences in this dataset, Figure \ref{fig:DETRAC-img} presents extracted frames of videos from Table \ref{tab:Res-2-DETRAC}. Our method performed best on the images from the first row while the worst results were obtained on the second row. The sunny weather condition seems to be a critical factor to improve the results, but we think the camera angle is playing an important role. Tracking while facing the traffic with a good elevation angle looks like the optimal setup to achieve best performance with our method. The same applies to IOU tracker indicating that detections could be of lesser quality in those setups. No sequence in a rainy environment are among the least successful sequences even if it was the category with the worst results globally, which is counterintuitive. We assume the explanation comes from the camera angles used; side perspectives and low elevation angles cause difficult tracking situations. This is logical since both factors increase the probability of occlusion. Examples supporting this hypothesis are in the MVI\_40761 sequence (Figure \ref{fig:DETRAC-img}e), where the bus is hiding the two most distant lanes for over half the frame and in the MVI\_39501 sequence (Figure \ref{fig:DETRAC-img}f) where even with a front view, tall vehicles are able to hide many others behind them.
        
    \subsection{2D MOT 2015}

        Table \ref{tab:interTestRes} compares our method with other approaches. The first two rows, \textit{HWDPL} \cite{ICIP2017Chen} and \textit{AMIR15} \cite{ICCV2017Sadeghian}, are to show the current performance of state-of-the-art methods that use multiple learned features, learned cost function and prediction in future frames. The last four entries include our method and three classic approaches using similar simple features as ours: \textit{DP NMS} (Min-cost flow) \cite{CVPR2011Pirsiavash}, \textit{TC ODAL} (Hungarian algorithm) \cite{CVPR2014Bae} and \textit{JPDA OP}, a JPDAF implementation from the MOT challenge website \cite{MOTChallenge2015}.
    
        \begin{table*}
            \centering
            \caption{Tracking results on the 2D MOT 2015 benchmark.}
            \begin{tabular}{lrrrrrrrr}
            \hline
            Method                             & \hspace{1em}$MOTA$ & \hspace{1em}$IDF1$ & $MT$   & $ML$   & $FP$  & $FN$  & $IDS$ & $Frag$ \\ \hline
            HWDPL \cite{ICIP2017Chen}          & 38.5   & 47.1   & 8.7\%  & 37.4\% & 4005  & 33203 & 586   & 1263 \\
            AMIR15 \cite{ICCV2017Sadeghian}    & 37.6   & 46.0   & 15.8\% & 26.8\% & 7933  & 29397 & 1026  & 2024 \\ \hline
            CP (ours)                                 & 17.0   & 13.3   & 2.5\%  & 58.4\% & 4872  & 43170 & 2973  & 3077 \\
            TC ODAL \cite{CVPR2014Bae}         & 15.1   & 0.0    & 3.2\%  & 55.8\% & 12970 & 38538 & 637   & 1716 \\
            DP NMS \cite{CVPR2011Pirsiavash}   & 14.5   & 19.7   & 6.0\%  & 40.8\% & 13171 & 34814 & 4537  & 3090 \\
            JPDA OP                            & 3.6    & 7.5    & 0.4\%  & 96.1\% & 1024  & 58189 & 29    & 119  \\ \hline
            \end{tabular}
            \label{tab:interTestRes}
        \end{table*}
    
         These results show that there are still many improvements to be made before we can match the performance of the state-of-the-art approaches, but it also shows that our method surpasses many classical approaches using similar features, even with our method using only a simple appearance model considering only ten color classes and the tendency to make many ID switch mistakes due to the CP model.
        
        The results are also significantly lower than those obtained on the UA-DETRAC benchmark. Tracking pedestrians instead of road vehicles comes with its own challenges. People move more slowly than motorized vehicles, but their general appearance is more variable because of the leg-and-arm complex movements required to walk. The 2D MOT 2015 dataset was created before the rise of deep learning, therefore the noise level contained in the set of public detections is higher. The frame rate is not the same for all sequences (from 10 to 30).
        
    \subsection{Ablation study}
    
        After comparing the method to others available, the next step was to check the performance of individual modules. We made this evaluation on a subset of the training set of UA-DETRAC, that includes 22 videos. Table \ref{tab:Res-ablation-DETRAC} reports the MOTA values obtained after three removals:
        \begin{enumerate}
            \item At first, the problem was converted to a strict constraint satisfaction problem (CSP). The \texttt{CostRegular} constraint is kept, but the trajectory costs are no longer optimized.
            \item Then, the \texttt{CostRegular} is completely removed. No appearance model is used at this point.
            \item Finally, the pre solving computations based on an VOT technique are deactivated.
        \end{enumerate}
    
        \begin{table}[hbt!]
            \centering
            \caption{Results on a subset of sequences from UA-DETRAC with different configurations of our method. Each removal is valid for the current row and all the row below.}
            \begin{tabular}{lrrrrr}
                \hline
                Setup       & \textit{MOTA} & \textit{IDS} & \textit{FN} & \textit{FP} & \textit{Exec (s)} \\ \hline
                All         & 80.23         & 11048        & 39831       & 8722        & 53747.70          \\
                CSP         & 80.01         & 11699        & 39835       & 8726        & 1180.25           \\
                No Regular  & 80.01         & 11710        & 39835       & 8726        & 1100.41           \\
                No Presolve & 78.88         & 12909        & 43626       & 7126        & 246.25            \\ \hline
            \end{tabular}
            \label{tab:Res-ablation-DETRAC}
        \end{table}
        
        Table \ref{tab:Res-ablation-DETRAC} confirms that each module helps improve the quality of the tracking, even if the increase is sometimes only small. Without the optimization process, the \texttt{CostRegular} is not removing many ID switch errors. The addition of an optimization objective allows a significant drop in the number of ID switch mistakes while maintaining approximately the same number of false positives and negatives. This indicates that a simple motion model is able to make mostly correct associations since forbidding associations based on vehicle appearances produces almost identical results.
        
        The pre solve process allows to eliminate a greater number of false negatives while not increasing too much the number of false positives, thus increasing the global MOTA value. It makes sense since this module is mainly creating new detections.
        
        The resolution time values were provided (in seconds) for each execution since it is an important factor to assess the quality of a CP model. All computations were done in a virtual machine running Ubuntu 19.10, either on a Intel i7-2600 processor from 2011, or a 2018 i9-9900k processor for the ablation study.
        
        Heavy computations are required to obtain the best results with this method. As MOT is a problem that often requires to be solved in real-time; video surveillance cameras are filming constantly, taking hours to solve a five-minute video sequence is not always a possibility. The ablation study presented in Table \ref{tab:Res-ablation-DETRAC} confirms that even without considering the appearance model and using single object tracking predictions to add detections, the tracking performance is not dropping that much. The number of frames in the sequences equal 35495, which means that real-time performance is achieved since all UA-DETRAC sequences are at 30 frames per second, even with the presolve process and the automaton activated. This is an advantage provided by our CP method; the problem is solved in a larger context with real time performance compared to the IOU tracker which is fast, but considers the association problem only two frames at a time.

\section{Remaining Challenges}
    
    \subsection{Computation times}
    
        MOT is a problem where the computation time is limited since completing the tracking in real-time is desired in many situations. Our approach using CP and all its components is not close to this level of efficiency. However, as we have shown, removing some component can allow achieving real-time with a small performance hit. Processing video sequences that contain often more than a thousand frames increases the number of possible combinations before obtaining a solution. The total number of objects throughout the complete sequence is also impacting the computation since it directly affects the domain of the main branching variables. As a solution to this, we applied the data association to blocks of consecutive frames. Recall that usually data association is performed only on two consecutive frames. Considering more frames allows taking better long-term decisions. It is therefore not mandatory to perform the data association with all the frames, if  at least, the associations are solved with bigger batches. Finally, the complexity of the appearance model (i.e. the number of possible appearance values) affects directly the time required to test each variable-value combination.
    
    \subsection{Appearance Model}

        The appearance model is important to indicate the presence of occlusions. The precision with which we describe detections will impact the tracking accuracy. One approach may be to use vectors to describe the object (histogram of oriented gradients, deep features). However, they are difficult to include in a CP model without doing clustering first which may reduce the precision of object descriptions. As said before, a complex model will also slow down the resolution process.

\section{Conclusion}
\label{concl}
    This paper introduced a novel data association method using constraint programming. More precisely the center position of objects and their colors are used as main features to investigate if CP can be a valid approach to solve this problem in the context of road user tracking. Our result show that CP can help improve tracking as we outperform the IOU tracker on UA-DETRAC, which is one of the best tracker on this road user MOT dataset. Our strategy of using object position and a simplified color model proves to be well suited for vehicle tracking.  There are still improvements required to be able to solve large instances and to capitalize better on road user appearance. Future work will include developing a more complex model that considers a larger number of features and optimizing the search strategy in order to improve run-time and performance on pedestrian tracking.

\section*{Acknowledgements}
\addcontentsline{toc}{section}{Acknowledgment}

    This research was supported by an IVADO fundamental research grant.

\bibliographystyle{IEEEtran}
\bibliography{IEEEabrv, ref}

\end{document}